\pdfoutput=1

\documentclass[11pt]{article}
\usepackage{authblk}
\usepackage{graphicx}
\usepackage[]{ACL2023}

\usepackage{times}
\usepackage{latexsym}

\usepackage[utf8]{inputenc}
\usepackage{CJKutf8}
\usepackage[T5]{fontenc}

\usepackage{microtype}

\usepackage{inconsolata}

\title{A Top-down Graph-based Tool for Modeling Classical Semantic Maps:\\ A Crosslinguistic Case Study of Supplementary Adverbs}

\author[1]{\textbf{Zhu Liu}}
\author[2]{\textbf{Cunliang Kong}}
\author[ \hspace{0.2em}1]{\textbf{Ying Liu}\thanks{\hspace{0.5em} Corresponding author}}
\author[2]{\textbf{Maosong Sun}}

\affil[1]{School of Humanities, Tsinghua University}
\affil[2]{Department of Computer Science and Technology, Tsinghua University}

\affil[ ]{\nolinkurl{{liuzhu22, yingliu,sms}@tsinghua.edu.cn}} 
\affil[ ]{\nolinkurl{cunliang.kong@outlook.com}}

\usepackage{booktabs}
\usepackage{tabularx}
\usepackage{caption}
\usepackage{amsfonts}
\usepackage{amsmath}
\usepackage{color,xcolor} 
\usepackage{xspace}
\usepackage{adjustbox}
\usepackage{multirow}

\usepackage{amssymb}
\usepackage{algpseudocode}
\usepackage{xcolor,colortbl}
\usepackage{algorithm}
\usepackage{dcolumn}

\makeatletter

\newcommand{\Rmnum}[1]{\mathrm{\expandafter\@slowromancap\romannumeral #1@}}
\makeatother

\begin{document}

\setcounter{page}{1}

\maketitle

\vspace{2cm} 

\begin{abstract}
Semantic map models (SMMs) construct a network-like conceptual space from cross-linguistic instances or forms, based on the connectivity hypothesis. This approach has been widely used to represent similarity and entailment relationships in cross-linguistic concept comparisons. However, most SMMs are manually built by human experts using bottom-up procedures, which are often labor-intensive and time-consuming. In this paper, we propose a novel graph-based algorithm that automatically generates conceptual spaces and SMMs in a top-down manner. The algorithm begins by creating a dense graph, which is subsequently pruned into maximum spanning trees, selected according to metrics we propose. These evaluation metrics include both intrinsic and extrinsic measures, considering factors such as network structure and the trade-off between precision and coverage. A case study on cross-linguistic supplementary adverbs demonstrates the effectiveness and efficiency of our model compared to human annotations and other automated methods. The tool is available at \url{https://github.com/RyanLiut/SemanticMapModel}. 

\end{abstract}

\section{Introduction}
\label{introduction}

\begin{figure}
    \centering
    \includegraphics[width=1.0\linewidth]{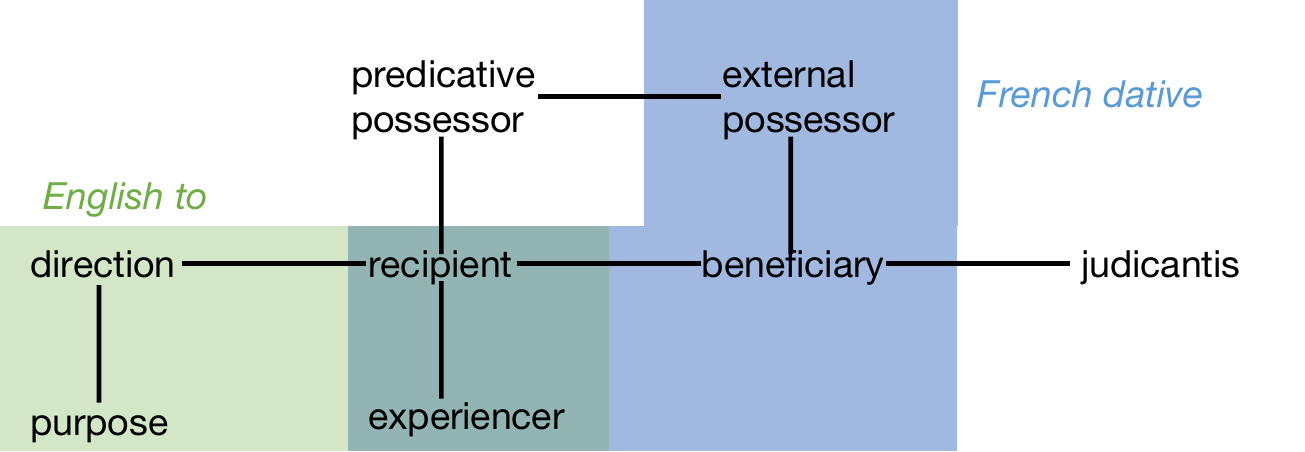}
    \caption{A semantic map of typical dative functions and the regions of English (green) to and French dative (blue).}
    \label{fig:smm-example}
\end{figure}

A linguistic form—such as a morpheme, word, or construction—can map to multiple related meanings or functions\footnote{We use the terms \textit{function} or \textit{concept} rather than \textit{sense} (conventional meanings) or \textit{use} (contextual meanings), as it is often difficult to distinguish between conventional, contextual, or even vague meanings, particularly in the case of function words or affixal categories~\cite{haspelmath2003geometry}.}, which, in turn, correspond to different cross-linguistic forms. To analyze this multifunctionality, linguists employ semantic map models (SMMs)~\cite{haspelmath2003geometry}. SMMs construct a network, known as conceptual space, where nodes represent distinct yet related functions, and edges reflect the semantic similarity between these functions. This structure adheres to the connectivity hypothesis~\cite{croft2001radical}, which asserts that functions shared by a single form should map onto a connected\footnote{Any two nodes in this region are connected either directly or indirectly through other nodes.} region in the conceptual space, thus preserving the limits of linguistic variation~\cite{greenberg1963universals}. An example illustrating typical dative functions and two forms is shown in Figure~\ref{fig:smm-example}.

\begin{figure*}[h]
    \centering
    \includegraphics[width=1.0\linewidth]{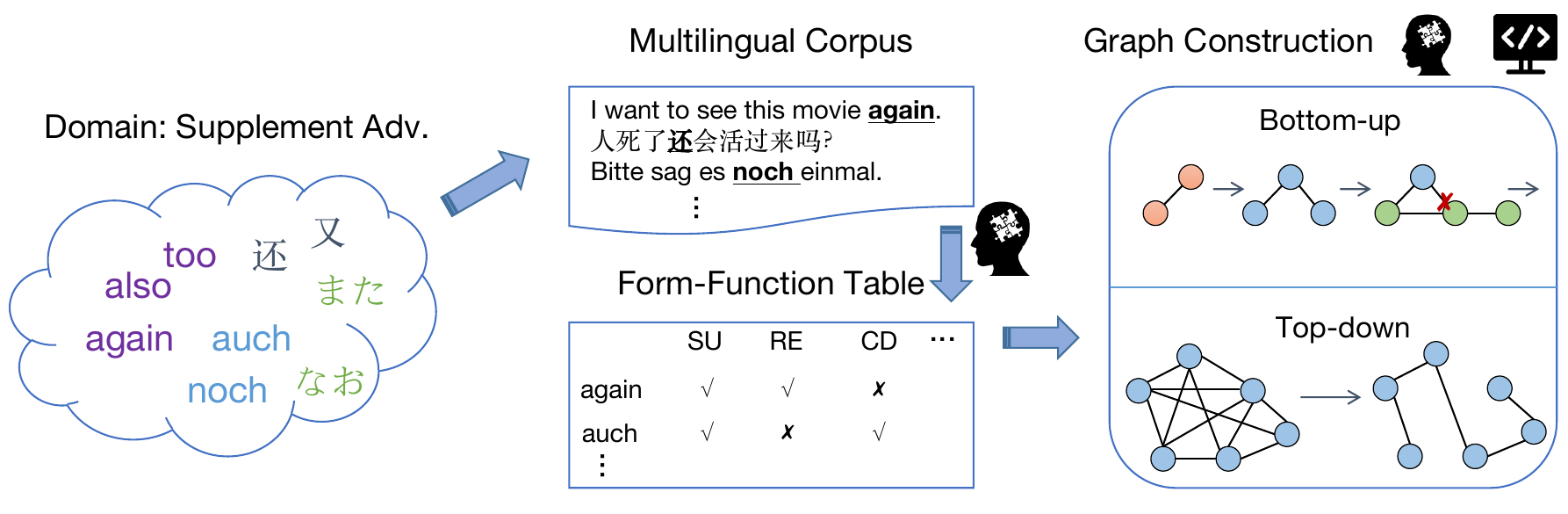}
    \caption{Three steps for constructing semnatic maps. First is to identify the semantic domain and related forms in multiple languages. Second, linguists analyze the form-function table based on the multilingual corpus. Third, a graph is constructed in either a bottom-up or top-down manner. Our method employs a top-down construction.}
    \label{fig:intro}
\end{figure*}

Constructing the conceptual space typically involves three steps. First, the semantic domain and the corresponding cross-linguistic forms are identified. Second, linguists analyze a form-function table, determining whether a given form can express a specific function based on a multilingual corpus. Third, edges are added and adjusted by either manual annotation or automated procedures, following bottom-up or top-down approaches guided by the connectivity hypothesis. This ensures that each subgraph—representing the multiple functions of a form (i.e., the rows in the form-function table)—remains connected. Typically, this process incrementally builds the graph, edge by edge, as new instances are encountered, reflecting a bottom-up approach. In contrast, our method introduces a top-down algorithm to improve efficiency. These steps are illustrated in Figure~\ref{fig:intro}.


However, this construction process has several limitations. First, manually adjusting the graph is both time-consuming and labor-intensive, especially when handling large datasets with a large number of languages and forms. Second, although some semi-automated methods have been proposed to assist~\cite{ma2015semantic_en}, they still adhere to a bottom-up approach, requiring substantial manual effort to determine which edges to retain or add. Third, human judgment introduces an element of subjectivity, particularly when deciding between two equally plausible connections.


To address these challenges, we propose a novel graph-based algorithm for constructing the conceptual space in a top-down manner. The algorithm begins by creating a dense graph, where the co-occurrence of functions (nodes) in the same form (colexification) serves as the weights for their corresponding edges. We then prune this graph by extracting maximum spanning trees, sorted in descending order by the total weight of their edges. The optimal trees are selected based on automated metrics that include coverage of linguistic instances and network topology. This process is considered top-down because it starts with a dense but suboptimal conceptual space and is subsequently pruned into a tree constrained by our proposed hypothesis. Therefore, it provides a global solution by considering all possible edges in the graph, rather than employing a bottom-up strategy that begins from a localized region formed by individual forms. A case study on supplementary adverbs demonstrates the model's competitive performance and efficiency compared to human-generated maps and other automated methods.

The key contributions of this paper are as follows:
\begin{itemize}
    \item We propose a top-down graph-based approach for constructing conceptual spaces and semantic map models (SMMs).
    \item We design a set of metrics to evaluate the quality of the resulting networks.
    \item A case study on supplementary adverbs demonstrates the efficiency and effectiveness of our proposed method.
    \item We develop a visualization tool based on this approach to assist typological linguists in studying multifunctionality across languages conveniently.
\end{itemize}



\section{Related Work}
\label{related work}

\begin{table*}[t]
    \centering
    \resizebox{1.0\textwidth}{!}{ 
    \begin{tabular}{ccc}
    \toprule
        Term & Notation & Description  \\
    \midrule
        form & $x \in \mathcal{X}$ & A linguistic unit, such as function words or constructions. \\
        function & $y \in \mathcal{Y}$ & Meanings or concepts represented as nodes. \\
        function-form table & $M$ & A binary indicator of whether a form expresses a specific function. \\
        degree of colexification & $w$ & Co-occurrence of functions, represented as weights. \\
        conceptual space & $\mathcal{G}(\{y \mid M(x,y) = 1\})$ & Represents the similarity between function nodes. \\
        semantic map for $x'$ & $\mathcal{G}(\{y \mid M(x',y) = 1\})$ & A connected boundary or region corresponding to a specific form $x'$. \\
    \bottomrule
    \end{tabular}
    }
    \caption{Notations and descriptions for key terms.}
    \label{tab:term}
\end{table*}


\paragraph{Semantic Map Models}
Semantic map models (SMMs) are designed to describe linguistic forms that convey different meanings, particularly within the realm of linguistic typology. These forms may include content words~\cite{guo2012adjectives, cysouw2007building}, function words~\cite{zhang2017semantic}, or constructions~\cite{malchukov2007ditransitive}. The distinctions between meanings can vary, encompassing word senses (for content words), grammatical functions (for function words), or even specific word forms~\cite{malchukov2007ditransitive}. SMMs are typically represented as graphs, where nodes correspond to individual functions and edges reflect the proximity between these functions. Classical SMMs~\cite{haspelmath2003geometry} connect nodes based on the connectivity hypothesis (see Subsection~\ref{subsec:space}). In contrast, second-generation models, such as those based on multi-dimensional scaling (MDS)~\cite{MDS}, introduce weighted edges determined by the frequency of co-occurrence between function nodes. Furthermore, a more general methodology for modeling conceptual spaces includes concept translation across languages~\cite{liu-etal-2023-crosslingual}, geometric representations derived from language models~\cite{Moullec2025} and other related approaches. Our method builds upon classical SMMs with the enhancement of assigning weights to edges.

Generally, linguists construct Semantic Map Models (SMMs) by manually adjusting edge connections to satisfy the necessary constraints. However, this process can be labor-intensive and time-consuming, particularly when dealing with large datasets. Several methods have been proposed to automate the map-building process. For instance, the author in \citet{ma2015semantic_en} developed an algorithm that determines when to add an edge for each form in a bottom-up manner. Nevertheless, their approach still requires human input to select the optimal connections for each form, thereby classifying it as a ``computer-assisted application.'' Additionally, their method only addresses the addition of edges and does not account for the removal of edges for new forms. The other two studies \citet{xiao2021review,chen2015revealing} concentrate on the second generation of SMMs, which is not the focus of our work.


\paragraph{Supplementary Adverbs}
Supplementary adverbs are universal across multiple languages, serving to modify the meaning of verbs, adjectives, or other adverbs in a sentence and thereby providing additional contextual information. These adverbs play a crucial role in communication by indicating time, degree, manner, or continuity. Researchers have conducted extensive studies on this subject, emphasizing their multifunctionality. Given this high degree of multifunctionality, the work in \cite{zhang2017semantic} distinguishes different functions in terms of semantic features, such as \textit{Bounded}, \textit{Sequenced}, etc. Linguists focus on words in their respective languages, while still adopting a cross-lingual perspective. For instance, the English adverb \textit{still} is often used to indicate the continuity or persistence of an action over time \cite{michaelis1996cross}. In German, \textit{noch} can denote an additional action or state, while \textit{schon} often implies that something has occurred earlier than expected \cite{konig1977temporal}. In Chinese, the adverb {\begin{CJK}{UTF8}{gbsn} 还 \end{CJK} is used to express the continuation or inclusion of an action, whereas {\begin{CJK}{UTF8}{gbsn} 有 \end{CJK} can indicate an additional aspect of an action~\cite{shen2001two, liu2000scalar}. Some research also illustrates the evolution of these related words in a diachronic~\cite{tong2004semantic} or synchronic~\cite{Guo2012Sync} context.
 

In this paper, we selected Supplementary Adverbs as a case study for two main reasons. First, they have been extensively studied, with significant consensus in the field. Second, the data scale encompassing a wide range of languages, forms, and functions is substantial, making it a particularly suitable and challenging scenario for constructing semantic maps.

\section{Preliminary}
In this section, we first present some notations, then introduce the conceptual space and the semantic map.

\subsection{Notations}

In this section, we first establish some notations, followed by an introduction to the conceptual space and the semantic map. We summarize these in Table~\ref{tab:term}.

\subsection{Nodes and Data Forms}
Given a core semantic domain $\mathcal{D}$, we select a candidate set of forms $\mathcal{X}$ and a corresponding set of potential functions $\mathcal{Y}$. Each form $x \in \mathcal{X}$ may be a word (primarily function words), a morpheme, or even a construction. We then examine a corpus to calculate the frequency $f(x,y)$ with which a specific form $x$ corresponds to the function $y \in \mathcal{Y}$. Consequently, we can create a binary\footnote{The binary setting is required in classical SMMs, although the value can also take other forms, such as the frequency of $x$ with respect to $y$ in a corpus, or even a probability for assigning $x$ to $y$. We leave this for further study.} function-form matrix $M \in \{0, 1\}^{m \times n}$, where the rows represent $m$ forms and the columns represent $n$ functions:
\begin{equation}
    M_{x,y} =
\begin{cases} 
1, & \text{if } f(x,y) \geq 1 \\
0, & \text{otherwise}.
\end{cases}
\end{equation}



\subsection{Conceptual Space}
\label{subsec:space}
A conceptual space represents the similarity between functions or concepts, adhering to the connectivity hypothesis \cite{croft2001radical}. This hypothesis posits that any form $x$ associated with different functions must correspond to a connected region. We can formalize the conceptual space as an undirected graph $\mathcal{G}$, where the nodes represent the function space $\mathcal{Y}$, and the edges must satisfy the following hypothesis.

\paragraph{Hypothesis 1} \label{hyp:1}
The subgraph $\mathcal{G}(\{y \mid M(x,y) = 1\})$ is connected, where $\mathcal{G}(\{y \mid M(x,y) = 1\})$ denotes the subgraph consisting of the function set associated with a form $x$.


However, Hypothesis 1 is relatively relaxed, allowing many graphs to satisfy the constraints. Therefore, we need to design metrics to evaluate the graph. We consider both intrinsic and extrinsic metrics, which are introduced in detail in Section~\ref{sec:in_metrics}.

\subsection{Semantic Map}
Given a conceptual space, a semantic map for a specific form $x'$ is defined as the connected region within the space, indicating the distribution of the functions that $x'$ encompasses. The map (as shown in Figure~\ref{fig:smm-example}) illustrates the proximity of these functions, implying their diachronic evolution and providing insights into language typology.

There are several variations of Semantic Map Models (SMMs). The classical model is based on Hypothesis 1 concerning connectivity. Another version \cite{MDS} incorporates weights for edges based on their frequency of co-occurrence between two functions in a corpus. This approach visualizes the nodes in a plot that preserves their similarity or distance according to each weight, utilizing tools such as MDS. However, this version discards Hypothesis 1, resulting in a graph that is less interpretable than the classical model. In our paper, we adopt the classical model while adding weights to the edges, aiming to leverage the advantages of both methods.

\section{Approach}
\label{sec:approach}
In this section, we first introduce our top-down graph-based algorithm for Semantic Map Models (SMMs). Following this, we design both intrinsic and extrinsic metrics.

\subsection{Top-down Graph-based Algorithm}
Considering the trade-off between coverage and precision, we adopt a top-down perspective: (1) to construct a dense graph (sufficient condition); and (2) to prune it into a sparser graph (necessary condition).

First, we construct a dense graph $G_0$, where each edge $e \langle y_i, y_j \rangle$ connecting two function nodes $y_i$ and $y_j$ has an associated weight $w(e)$, referred to as the unnormalized degree of association~\cite{guo2012concepts}. This weight is calculated as follows:
\begin{equation}
    w(e \langle y_i, y_j \rangle) = M(:, y_i) \cdot M(:, y_j),
\label{equ:weight}
\end{equation}
where $M(:, y_i)$ represents the $i$th column of the matrix $M$. Thus, $w(e)$ indicates the number of co-occurrences for the two functions\footnote{An alternative version~\cite{guo2012concepts} normalizes $w$ by the occurrence frequency of either individual $y_i$ or $y_j$.}.

Next, we remove some edges from the complete graph $G_0$. We introduce three constraints into the topology of the conceptual space, as induced by Hypothesis 1:
\begin{enumerate}
    \item The first constraint is that the final graph $G$ should be connected. This is crucial because if $G$ contains multiple unconnected components, the data can be partitioned into corresponding parts for separate analysis.
    \item The second constraint is acyclicity~\cite{zhang2015semantic}; cycles among the function nodes do not adequately represent entailment relations and fail to predict some wrong forms~\footnote{For example, a sequential connection of A, B, and C can predict a form with both A and C as impossible due to the disconnected region, while a cycle cannot.} and should be avoided whenever possible, unless there are no other solutions available to cover the data. Haspelmath~\cite{haspelmath2003geometry} refers to the loop as a ``vacuous map'' because it has no predictive power.
    \item The third constraint pertains to the size of $G$; specifically, the summed weights of all edges should be maximized. This is important because these edges represent the most similar functions, which should be retained as much as possible.
\end{enumerate}

These three additional constraints effectively create a maximum spanning tree using graph theory. The revised Hypothesis 2, derived from Hypothesis 1, can be formalized as follows:

\paragraph{Hypothesis 2} 
$\mathcal{G}(\{y \mid M(x,y) = 1\})$ is a maximum spanning tree, where the weights $w(e)$ on each edge $e$ are calculated using Equation~\ref{equ:weight}.

In practice, we obtain $n$ spanning trees sorted by their size. Each tree is then evaluated using intrinsic or extrinsic metrics $m$. Algorithm~\ref{alg:top-down} illustrates this process. We utilize the Python package \textit{networkx} to compute the maximum spanning trees~\footnote{\url{https://networkx.org/documentation/stable/reference/algorithms/generated/networkx.algorithms.tree.mst.SpanningTreeIterator.html}} sorted by weights, i.e., using the function $MaxSpanningTrees()$. Additionally, we assess the model, as described in the $Evaluate()$ function in the next subsection.

\begin{algorithm}
\caption{Top-down Graph-based SMMs}
\textbf{Input:} $M$, number of candidate trees $n$, metrics $m$\\
\textbf{Output:} $G$, the set of conceptual spaces
\begin{algorithmic}[1]
\State $G \gets \emptyset$
\State Construct $G_0$ from $M$ with weights by Equation~\ref{equ:weight}
\State $P \gets MaxSpanningTrees(G_0)$
\For{each tree $t$ in the top $n$ trees in $P$}
    \If {$Evaluate(t)$ satisfies $m$}
    \State $G \gets G \cup \{t\}$
    \EndIf
\EndFor
\State \textbf{return} $G$
\end{algorithmic}
\label{alg:top-down}
\end{algorithm}

\begin{figure}
    \centering
    \includegraphics[width=1.0\linewidth]{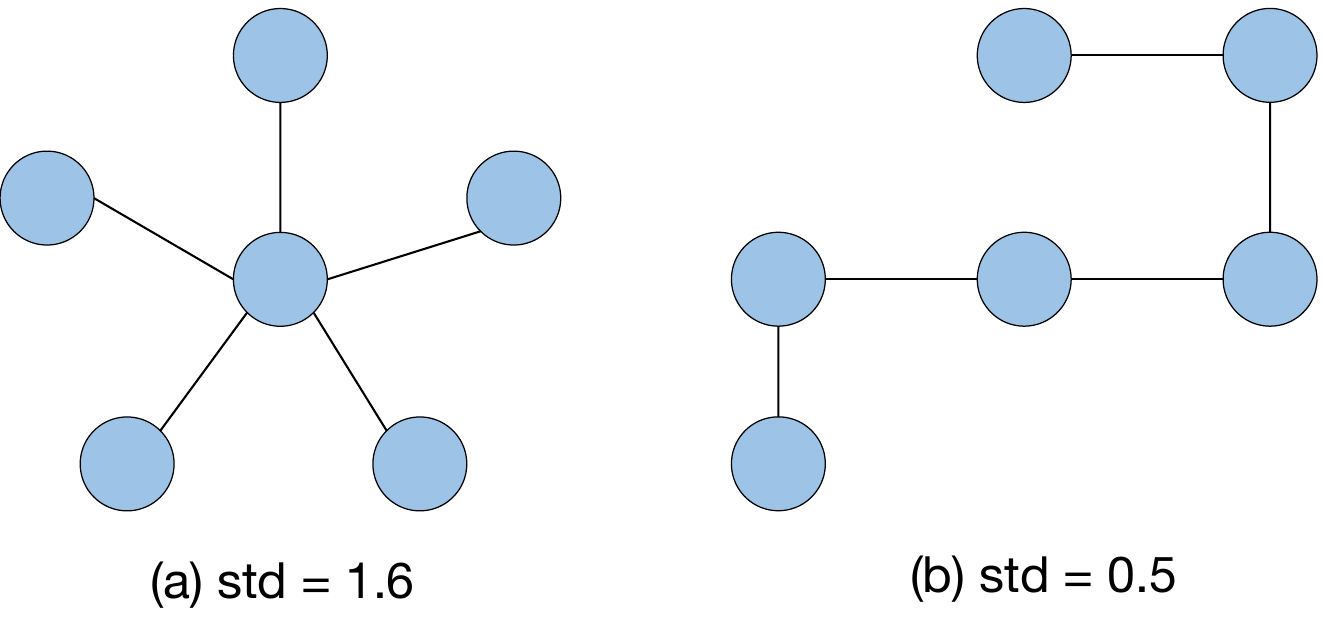}
    \caption{Network topology with different standard deviations of degrees. The left shows a star-like graph with a central node connecting other nodes, while the right shows an averaged connectivity for every node.}
    \label{fig:typology}
\end{figure}

\subsection{Intrinsic Metrics} \label{sec:in_metrics}
For a candidate graph $G$, we evaluate it based on recall, precision, and topology. Recall measures the proportion of forms that satisfy Hypothesis 1 relative to the total number of candidate forms. In contrast, precision shares the same numerator as recall, but its denominator is the number of all possible connected components given the constructed graph. Recall reflects the coverage of the actual samples, while precision ensures that the graph maintains limited predictive power. The trade-off between precision and recall restricts us in finding an optimal graph that is both sufficient (recall) and necessary (precision). For instance, a complete graph serves as a trivial solution that satisfies Hypothesis 1 with a recall of 1, but it possesses very low predictive power, as it tends to predict many noisy forms.

We also propose a new metric for the topology of the network. Specifically, we compute the standard deviation (Div\_D) of all the degrees, which is the number of edges connected to a specific node. A lower standard deviation is desirable because it indicates that the edges from the nodes are more ``averaged,'' rather than exhibiting a star-like topology, which offers \textcolor{black}{less interpretability for the entailment relationship of functions}. For instance, Croft~\cite{croft2003typology} employs a Markov chain-style representation (Figure~\ref{fig:typology}b) rather than a star-like network (Figure~\ref{fig:typology}a) to depict the animacy hierarchy across multiple languages.
Additionally, we use the size of the graph, i.e., the summed weights of all the connected edges, to indicate the important relationships that are retained. Table~\ref{tab:descirption} presents brief descriptions of the different metrics.

\begin{table}[h]
    \centering
    \small
    \begin{tabular}{ccc}
    \toprule
        Metric & Description & Trend \\
    \midrule
        Size & Summed weights of edges & $\uparrow$ \\
        Recall & Coverage rate of instances & $\uparrow$ \\
        Precision & Accuracy of predicted instances & $\uparrow$ \\
        Div\_D & Standard deviation of degrees & $\downarrow$ \\
    \midrule
    \midrule
        Acc & Matched rate compared to GT & $\uparrow$ \\
        \bottomrule
    \end{tabular}
    \caption{Different metrics for evaluating the conceptual space. The trend shows the optimal direction for a better network.}
    \label{tab:descirption}
\end{table}

\begin{table*}[t]
    \centering
    \scriptsize
    \setlength{\tabcolsep}{6.5pt}
    \begin{tabular}{c|c|cccccccccccccccccc}
    \toprule
         L & G & AF & SU & RE & CO & GD & DE & IS & CD & DC & PT & SC & WH & SE & SC & IC & UE & BL & DS  \\
        \midrule
         \multirow{4}{*}{ZH} & {\begin{CJK}{UTF8}{gbsn} 还 \end{CJK}} & 0 & 1 & 1 & 1 & 1 & 1 & 1 & 1 & 1 & 1 & 0 & 0 & 0 & 0 & 0 & 1 & 1 & 0 \\
            & {\begin{CJK}{UTF8}{gbsn} 又 \end{CJK}} & 0 & 1 & 1 & 0 & 0 & 0 & 1 & 0 & 0 & 0 & 0 & 0 & 0 & 0 & 1 & 0 & 0 & 1 \\
         & {\begin{CJK}{UTF8}{gbsn} 也 \end{CJK}} & 1 & 1 & 0 & 0 & 0 & 0 & 0 & 1 & 1 & 1 & 1 & 0 & 0 & 1 & 0 & 0 & 1 & 0  \\
         & {\begin{CJK}{UTF8}{gbsn} 在 \end{CJK}} & 0 & 1 & 1 & 1 & 1 & 0 & 1 & 0 & 0 & 0 & 0 & 1 & 1 & 0 & 0 & 0 & 0 & 0 \\
         \midrule
         \multirow{2}{*}{BO} & ra & 1 & 1 & 0 & 0 & 0 & 0 & 0 & 1 & 0 & 1 & 1 & 0 & 0 & 0 & 0 & 0 & 0 & 0 \\
             & tarong & 0 & 0 & 1 & 1 & 0 & 0 & 0 & 1 & 0 & 0 & 0 & 0 & 0 & 0 & 0 & 0 & 0 & 1\\
             \midrule
        \multirow{4}{*}{EN} & also & 1 & 1 & 0 & 0 & 0 & 0 & 0 & 0 & 0 & 0 & 0 & 0 & 0 & 0 & 0 & 0 & 0 & 0\\
         & too & 1 & 1 & 0 & 0 & 0 & 0 & 0 & 0 & 0 & 0 & 0 & 0 & 0 & 0 & 0 & 0 & 0 & 0 \\
         & again & 0 & 1 & 1 & 0 & 0 & 0 & 1 & 0 & 0 & 0 & 0 & 0 & 0 & 0 & 0 & 0 & 0 & 1 \\
         & still & 0 & 0 & 0 & 1 & 1 & 1 & 0 & 1 & 1 & 0 & 0 & 0 & 0 & 0 & 0 & 0 & 1 & 0 \\
         \midrule
         \multirow{2}{*}{DE} & auch & 1 & 1 & 0 & 0 & 0 & 0 & 0 & 1 & 0 & 1 & 1 & 0 & 0 & 1 & 0 & 0 & 0 & 0\\
         & noch & 0 & 1 & 1 & 1 & 1 & 1 & 0 & 1 & 1 & 0 & 0 & 1 & 0 & 0 & 0 & 1 & 1 & 0 \\
         \midrule
         \multirow{2}{*}{FR} & aussi & 1 & 1 & 0 & 0 & 0 & 0 & 0 & 0 & 0 & 0 & 0 & 1 & 0 & 0 & 0 & 0 & 0 & 0 \\
         & encore & 0 & 1 & 1 & 1 & 1 & 0 & 0 & 0 & 0 & 0 & 0 & 0 & 0 & 0 & 0 & 0 & 0 & 0\\
         \midrule
         \multirow{2}{*}{RU} & tbzhe  & 1 & 1 & 0 & 0 & 0 & 0 & 0 & 1 & 0 & 0 & 0 & 0 & 0 & 0 & 0 & 0 & 0 & 0 \\
         & opyat & 0 & 1 & 1 & 0 & 0 & 0 & 0 & 0 & 0 & 0 & 0 & 0 & 0 & 0 & 0 & 0 & 0 & 0\\ 
         \midrule
         \multirow{3}{*}{JA} & \begin{CJK}{UTF8}{min}   も \end{CJK} & 1 & 1 & 0 & 0 & 0 & 0 & 0 & 1 & 0 & 1 & 0 & 0 & 0 & 0 & 0 & 0 & 0 & 0 \\
         & \begin{CJK}{UTF8}{min}  また \end{CJK} & 0 & 1 & 1 & 0 & 0 & 0 & 0 & 0 & 0 & 0 & 0 & 0 & 0 & 0 & 0 & 0 & 0 & 0 \\
         & \begin{CJK}{UTF8}{min} なお \end{CJK} & 0 & 0 & 0 & 1 & 1 & 0 & 0 & 0 & 0 & 0 & 0 & 0 & 0 & 0 & 0 & 0 & 0 & 0 \\
         \midrule
         \multirow{5}{*}{KO} & {\begin{CJK}{UTF8}{mj} 도 \end{CJK}} & 1 & 1 & 0 & 0 & 0 & 0 & 0 & 1 & 1 & 1 & 0 & 0 & 0 & 0 & 0 & 0 & 0 & 0\\
         & {\begin{CJK}{UTF8}{mj} 더 \end{CJK}} & 0 & 0 & 0 & 0 & 1 & 0 & 0 & 0 & 0 & 0 & 0 & 0 & 0 & 0 & 0 & 0 & 0 & 0 \\
         & {\begin{CJK}{UTF8}{mj} 또 \end{CJK}} & 0 & 1 & 1 & 0 & 0 & 0 & 0 & 0 & 0 & 0 & 0 & 0 & 0 & 0 & 1 & 0 & 0 & 1\\
         & {\begin{CJK}{UTF8}{mj} 다시 \end{CJK}}  & 0 & 0 & 1 & 0 & 0 & 0 & 1 & 0 & 0 & 0 & 0 & 0 & 0 & 0 & 0 & 0 & 0 & 0 \\
         & {\begin{CJK}{UTF8}{mj} 아직 \end{CJK}} & 0 & 0 & 0 & 1 & 0 & 1 & 0 & 0 & 0 & 0 & 0 & 0 & 0 & 0 & 0 & 0 & 0 & 0\\
         \midrule
        \multirow{4}{*}{VI} & cũng & 1 & 0 & 0 & 0 & 0 & 0 & 0 & 1 & 1 & 1 & 0 & 0 & 0 & 0 & 0 & 1 & 1 & 0\\

        
        & nữa & 0 & 1 & 1 & 1 & 0 & 0 & 0 & 0 & 0 & 0 & 0 & 0 & 0 & 0 & 0 & 0 & 0 & 0\\
        & còn & 0 & 0 & 1 & 1 & 1 & 0 & 0 & 1 & 0 & 1 & 0 & 0 & 0 & 0 & 0 & 0 & 0 & 0\\
        & lại & 0 & 1 & 1 & 0 & 0 & 0 & 1 & 0 & 0 & 0 & 0 & 0 & 0 & 0 & 1 & 0 & 0 & 0\\
        \bottomrule
         
    \end{tabular}
    \caption{Form-function table for the Supplement-related semantic domain. Here, ``L'' represents languages and ``G'' denotes grams. Abbreviations for languages and functions are detailed in Tables~\ref{tab:languages} and~\ref{tab:functions}. A value of ``1'' indicates that the gram corresponds to the function in at least one sentence.}

    \label{tab:gram-func}
\end{table*}

\subsection{Extrinsic Metrics}
We also evaluate the network extrinsically, assuming we have the ground-truth semantic space constructed by linguistic experts. We calculate the accuracy (acc) as the ratio of matched edges to the total number of edges. Specifically, we represent a graph with an adjacency matrix $T$, using $T_p$ for the candidate graph and $T_g$ for the ground truth (GT). The accuracy is calculated as follows:
\begin{equation}
    \text{acc} = \frac{\sum_{i,j} \mathbf{1}_{T_p(i,j) = T_g(i,j)}}{n^2},
\end{equation}
where $n$ is the number of functions. 

We emphasize that several baselines are necessary to determine the lower bound of accuracy. The first baseline is a tree (LT) in which the edges do not overlap with the GT. The lower bound can be calculated as:
\begin{equation}
    LB_{LT} = \frac{n^2 - 4 \times (n-1)}{n^2}.
\end{equation}

The second baseline is a complete graph, which is the opposite of the first case: only the edges in the ground truth (GT) are correctly chosen. The lower bound for this scenario can be expressed as:
\begin{equation}
    LB_{C} = \frac{4 \times (n-1) + n}{n^2}.
\end{equation}

\section{Case Study}
In this section, we present a case study that utilizes Semantic Map Models (SMMs) to analyze adverbs related to "Supplement"~\cite{guo2010semantic, Guo2012Sync, zhang2015semantic}. The author collected 28 forms (function words) across nine languages that exhibit the "Supplement" function along with 17 other related functions~\footnote{We excluded the "Conjunction" function, as no words were associated with it.}. Based on whether a form in a corpus corresponds to a given function, the author constructed the form-function table \( M \), as shown in Table~\ref{tab:gram-func}. The full names of the languages and functions are provided in Tables~\ref{tab:languages} and~\ref{tab:functions}. Then he built the semantic map which we leverage as ground truth, as shown by the purple solid lines in Figure~\ref{fig:CS0}. We emphasize that this only serves as an acceptable, albeit not strictly standardized, solution. Other researchers~\cite{zhang2017semantic} may propose sightly constructions due to variations in the corpus or network-building methodology.

\begin{table}[h]
    \centering
    \setlength{\tabcolsep}{6.0pt}
    \begin{tabular}{cc}
    \toprule
    Abbr & Full \\
    \midrule
         ZH & Chinese  \\
         BO & Tibetan \\
         EN & English \\
         DE & German \\
         FR & French \\
         RU & Russian \\
         JA & Japanese\\
         KO & Korean \\
         VI & Vietnamese \\
        \bottomrule
    \end{tabular}
    \caption{ISO 639 abbreviation codes and full names for languages used in the case study.}
    \label{tab:languages}
\end{table}

\begin{table}[]
    \centering
    \setlength{\tabcolsep}{8pt} 
    \begin{tabular}{cc}
    \toprule
    Abbr & Full \\
    \midrule
         AF & Additive Focus  \\
         SU & Supplement \\
         RE & Repetition \\
         CO & Continuation \\
         GD & Greater Degree \\
         DE & Decrement \\
         CD & Condition \\
         DC & Discretional Condition \\
         PT & Polarity Trigger \\ 
         SC & Serious Condition \\
         WH & Whatever \\
         SE & Sequence \\
         SD & Sequential Coordinator \\
         IC & Inconsistency \\ 
         UE & Unexpectedness \\
         BL & Bottom Line \\
         DS & Discourse Continuation \\
    \bottomrule
    \end{tabular}
    \caption{Abbreviation and full names for functions.}
    \label{tab:functions}
\end{table}

\paragraph{Qualitative Analysis} We generated the conceptual space using the algorithm described earlier, with the graph exhibiting the largest weight displayed in Figure~\ref{fig:CS0} (purple solid lines), alongside the ground truth (GT) graph (black dashed lines). Compared to the GT network, our method reveals several distinctive features: (1) Each edge is assigned a weight that reflects the degree of connectivity between two functions; (2) Our network is acyclic, as stated in Hypothesis 2, while the GT includes cycles to maintain connectivity. We argue that introducing cycles should be avoided when possible due to the lack of interpretability~\cite{haspelmath2003geometry}, deferring this decision to human experts; (3) The basic topology of our network is similar to the GT; for instance, functions like "SU," "CD," and "RE" serve as critical points with multiple connecting edges.

\paragraph{Quantitative Analysis} Before evaluating the graphs sorted by size in descending order, we first identify the starting index for different graph sizes, as shown in Table~\ref{tab:boi}. The data indicate that the number of graphs at each size increases exponentially as the size grows, suggesting a wide variety of candidates and highlighting the importance of the evaluation metrics. Subsequently, we evaluate the top five graphs sorted by size, using a step size of 10K due to varying edge sizes, with results summarized in Table~\ref{tab:compare}. We also provide two baselines: one for the complete graph (denoted as C) and another for a tree with no overlap with the GT (denoted as LT). For comparison, the evaluation of the ground truth (denoted as GT) serves as the upper bound.

The results demonstrate that our method uncovers numerous comparable conceptual spaces, with recall exceeding 85\% and accuracy surpassing 90\%. It is important to note that precision is relatively low for each graph due to the high number of connected subgraphs. However, we identified instances where precision exceeds that of the ground truth (GT) while maintaining a similar level of recall (e.g., index 4). This finding suggests that linguists can utilize our algorithm to generate a set of comparable candidates and subsequently adjust the graph according to their expertise. We observed four forms in the largest network of Figure~\ref{fig:CS0} that violate connectivity hypothesis 1. These forms are \textit{taro} (RE-CO-CD-DS), \textit{still} (CO-GD-DE-CD-BL), \textit{cũng} (AF-CD-DC-PT-UE-BL), and \textit{còn} (RE-CO-GD-CD-PT). Notably, all share the common function CD, indicating a cycle involving this function, as shown by the GT.

\begin{table}[]
    \centering
    \begin{tabular}{cccc}
    \toprule
         Size & 90 & 89  & 88\\
         \midrule
         Begin of Index (boi) & 0 & 1,440 & 21,744 \\
         \bottomrule
    \end{tabular}
    \caption{Beginning of index (boi) corresponding to different sizes for a tree.}
    \label{tab:boi}
\end{table}

\begin{figure*}
    \centering
    \includegraphics[width=0.85\linewidth]{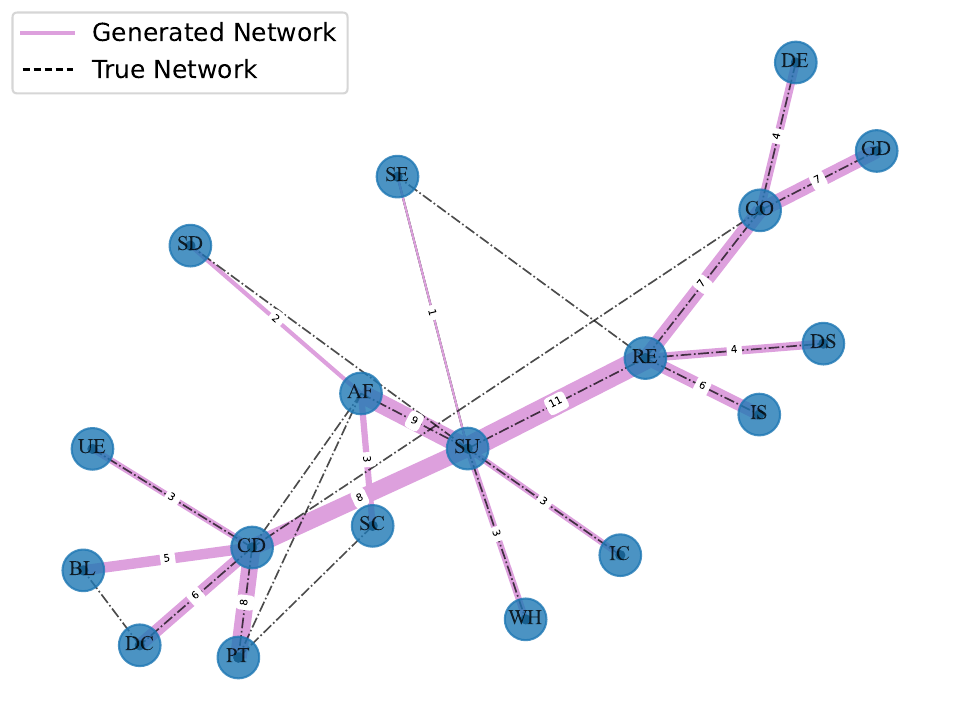}
    \caption{Tree of conceptual space with the largest size. The pink connections represent the network generated by our method, while the black dashed line indicates the ground truth as labeled by an expert. Numbers on the edge indicate the number of co-occurrences in a same word for the corresponding functions.}
    \label{fig:CS0}
\end{figure*}

\begin{table}[]
    \centering
    \small
    \begin{tabular}{cccccc}
    \toprule
         Index & Size$\uparrow$ & Recall$\uparrow$ & Precision$\uparrow$ & Accuracy$\uparrow$  \\
    \midrule
         C & 286 & 1 & 0 & 50.0 \\ 
         LT & - & - & - & 79.0 \\
         GT & 91 & 1 & 0.20 & 1 \\
    \midrule
         0 & 90 & 85.7 & 0.17 & 92.6 \\
         1 & 89 & 82.1 & 0.21 & 91.4 \\
         2 & 89 & 82.1 & 0.44 & 90.1 \\
         3 & 88 & 82.1 & 0.34 & 91.4 \\
         4 & 88 & 78.6 & 0.50 & 88.9 \\
         \bottomrule
    \end{tabular}
    \caption{Evaluation of our generated graphs and baselines (denoted as complete graph C and ground truth GT). The index represents the first N maximum spanning trees, scaled by 10,000.}
    \label{tab:compare}
\end{table}

\begin{table}[h]
    \centering
    \begin{tabular}{c@{\hskip 3pt} >{\raggedleft\arraybackslash}p{1.5cm}@{\hskip 6pt} >{\raggedleft\arraybackslash}p{1.5cm}}
    \toprule
         Round & RG\_1 & RG\_2  \\
    \midrule
         1 & -17.8 & -22.1 \\
         2 & -21.9 & -22.4 \\
         3 & -20.5 & -19.2 \\
         4 & -23.8 & -21.7 \\
         5 & -23.1 & -24.1 \\
    \midrule
         Mean & -21.4 & -21.9 \\
         Std. Dev. & 2.13 & 1.58 \\
    \bottomrule
    \end{tabular}
    \caption{Pearson correlation between Div\_D (diversity of degrees) and accuracy across five rounds. The mean and standard deviation for each round are also provided.}
    \label{tab:corr}
\end{table}

\paragraph{Metrics Analysis} We evaluate the effectiveness of our proposed metric, Div\_D. Given that tree structures typically exhibit less diversity in node degrees due to their sparse edges, we do not restrict our analysis to tree-based graphs. Instead, we generate graphs randomly in two ways: the first method produces graphs with entirely random edges, referred to as "RG\_1," while the second method (``RG\_2'') assigns edges with a probability proportional to the original weights, favoring higher-weighted edges. We sample 1,000 graphs per round over five rounds. In each round, we compute the Pearson correlation between Div\_D and accuracy, as shown in Table~\ref{tab:corr}. Both cases exhibit a moderate negative correlation, indicating that star-like topologies, characterized by lower degree diversity, are less desirable for an ideal conceptual space.

\section{Conclusion}
In this paper, we propose a graph-based algorithm to generate a conceptual space and semantic map in a top-down manner. This approach is realized by instantiating the conventional connectivity hypothesis into a greedy maximum spanning tree, where the weights are determined by the number of co-occurrences. To select the optimal candidate, we introduce a novel metric—diversity of degrees—based on graph topology, in addition to classical metrics such as precision, recall, and accuracy against ground truth. Through a case study on supplementary adverbs, we demonstrate the effectiveness and efficiency of our approach. Our visualization tool provides linguists with a valuable reference that can be manually refined further, while offering the computational community a graph-based framework for cross-linguistic semantics. This framework can be integrated with models like graph neural networks~\cite{4700287}, or approaches like conceptual space modeling~\cite{gardenfors2004conceptual}. 

\section{Limitations}
Our current version is preliminary and has several limitations. First, we do not account for the frequency with which a form is associated with a specific function in a corpus, an important factor in second-generation SMMs, such as MDS-based models. This may induce some noisy data. In future work, this frequency could be integrated into the edge weights. Second, we have not addressed the subjectivity and uncertainty involved in assigning functions to forms. The grams studied in SMMs, particularly function words like conjunctions, often exhibit highly uncertain and overlapping semantic spaces. We plan to mitigate this issue by employing soft labels instead of deterministic ones when populating the data matrix. Third, we have not incorporated temporal information into the evaluation of functions. While this aspect is more relevant to linguistic analysis, we aim to explore directional relationships in an automated manner in future research. Moving forward, we will validate our model with additional case studies across different languages and time periods. Last but not the least, this area suffers from a lack of comprehensive data and systematic evaluation metrics. Continued efforts are needed to expand the dataset and refine the evaluation process.

\section{Ethics Statement}
We do not foresee any immediate negative ethical consequences arising from our research.

\section{Acknowledgements}
The authors thank anonymous reviewers for their valuable comments and constructive feedback on the manuscript. We also thank Rui Guo, Bing Qiu, Minghu Jiang, Pengyuan Wang and Lei Dai for their valuable discussions. This work is supported by the 2018 National Major Program of Philosophy and Social Science Fund “Analyses and Researches of Classic Texts of Classical Literature Based on Big Data Technology” (18ZDA238), the National Natural Science Foundation of China (No. 62236011), Research on the Long-Term Goals and Development Plan for National Language and Script Work by 2035 (ZDA145-6) and Institute Guo Qiang at Tsinghua University.



\bibliography{custom}
\bibliographystyle{acl_natbib}
\clearpage









\end{document}